%% file: main.tex
\documentclass[runningheads]{llncs}
\usepackage{graphicx}
\usepackage{cite}
\usepackage{graphicx}
\usepackage{booktabs}
\usepackage{multirow}
\usepackage{subfigure}
\input{macro}

\usepackage{amsmath,amssymb} 
\usepackage{color}
\usepackage{subfigure}
\usepackage[width=122mm,left=12mm,paperwidth=146mm,height=193mm,top=12mm,paperheight=217mm]{geometry}

\graphicspath{ {figures/} }
\makeatletter
\newcommand{\printfnsymbol}[1]{%
  \textsuperscript{\@fnsymbol{#1}}%
}
\makeatother

\begin{document}
\title{BiSeNet: Bilateral Segmentation Network for Real-time Semantic Segmentation} 

\titlerunning{BiSeNet}
%
\author{Changqian Yu\thanks{Equal Contribution}\inst{1}\orcidID{0000-0002-4488-4157} \and
Jingbo Wang\printfnsymbol{1}\inst{2}\orcidID{0000-0001-9700-6262} \and
Chao Peng\inst{3}\orcidID{0000-0003-4069-4775} \and Changxin Gao\thanks{Corresponding Author}\inst{1}\orcidID{0000-0003-2736-3920} \and Gang Yu\inst{3}\orcidID{0000-0001-5570-2710} \and Nong Sang\inst{1}\orcidID{0000-0002-9167-1496}}
%
\authorrunning{C. Yu et al.}
%

\institute{National Key Laboratory of Science and Technology on Multispectral Information Processing,School of Automation,Huazhong University of Science \& Technology,China\\
\email{\{changqian\_yu,cgao,nsang\}@hust.edu.cn} \and
Key Laboratory of Machine Perception, Peking University, China \\
\email{wangjingbo1219@pku.edu.cn} \and
Megvii Inc. (Face++), China \\
\email{\{pengchao,yugang\}@megvii.com}}
\maketitle              
\begin{abstract}
Semantic segmentation requires both rich spatial information and sizeable receptive field. However, modern approaches usually compromise spatial resolution to achieve real-time inference speed, which leads to poor performance. In this paper, we address this dilemma with a novel Bilateral Segmentation Network (BiSeNet). We first design a Spatial Path with a small stride to preserve the spatial information and generate high-resolution features. Meanwhile, a Context Path with a fast downsampling strategy is employed to obtain sufficient receptive field. On top of the two paths, we introduce a new Feature Fusion Module to combine features efficiently. The proposed architecture makes a right balance between the speed and segmentation performance on Cityscapes, CamVid, and COCO-Stuff datasets. Specifically, for a 2048$\times$1024 input, we achieve 68.4\% Mean IOU on the Cityscapes test dataset with speed of 105 FPS on one NVIDIA Titan XP card, which is significantly faster than the existing methods with comparable performance.

\end{abstract}
\section{Introduction}
	\label{sec:intro}
	The research of semantic segmentation, which amounts to assign semantic labels to each pixel, is a fundamental task in computer vision. It can be broadly applied to the fields of augmented reality devices, autonomous driving, and video surveillance. These applications have a high demand for efficient inference speed for fast interaction or response. 
	
	Recently, the algorithms~\cite{Badrinarayanan-PAMI-SegNet-2017, Paszke-Arxiv-ENet-2016, Zhao-Arxiv-ICNet-2017, Li-CVPR-DLC-2017} of real-time semantic segmentation have shown that there are mainly three approaches to accelerate the model. 1) \cite{Zhao-Arxiv-ICNet-2017, Wu-Arxiv-Sparsity-2017} try to restrict the input size to reduce the computation complexity by cropping or resizing. Though the method is simple and effective, the loss of spatial details corrupts the predication especially around boundaries, leading to the accuracy decrease on both metrics and visualization. 2) Instead of resizing the input image, some works prune the channels of the network to boost the inference speed~\cite{Badrinarayanan-PAMI-SegNet-2017, Paszke-Arxiv-ENet-2016,Chollet-CVPR-Xception-2017}, especially in the early stages of the base model. However, it weakens the spatial capacity.  
	\begin{figure}[t]
		\centering
		\subfigure[Input and model]
		{
			\begin{minipage}{0.315\textwidth}
				\label{fig:input-model}
				\centering
				\includegraphics[width=1.0\linewidth]{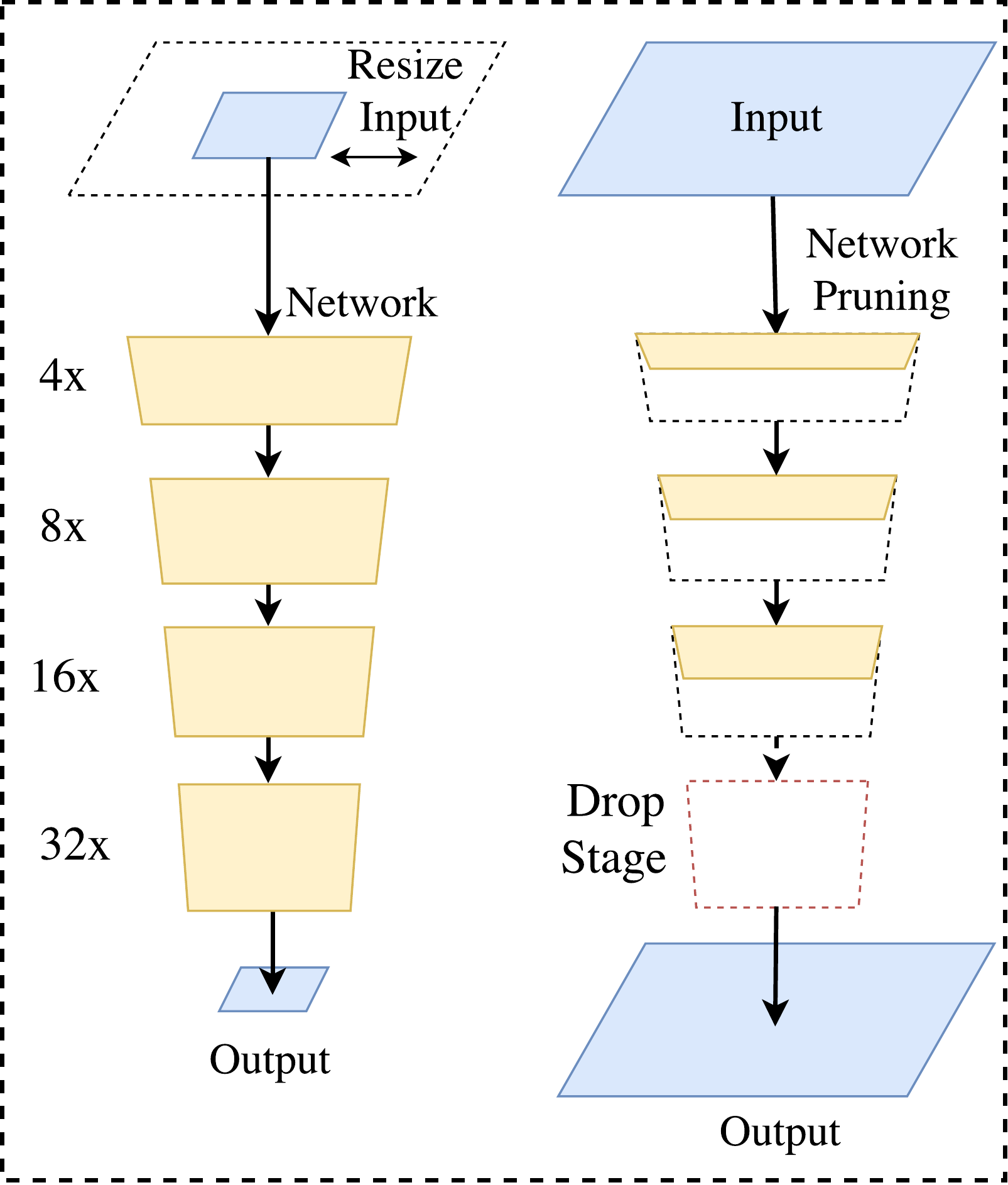}
			\end{minipage}
		}
		\hspace{-2ex}
		\subfigure[U-shape]
		{
			\begin{minipage}{0.315\textwidth}
				\label{fig:u-shape}
				\centering
				\includegraphics[width=1.0\linewidth]{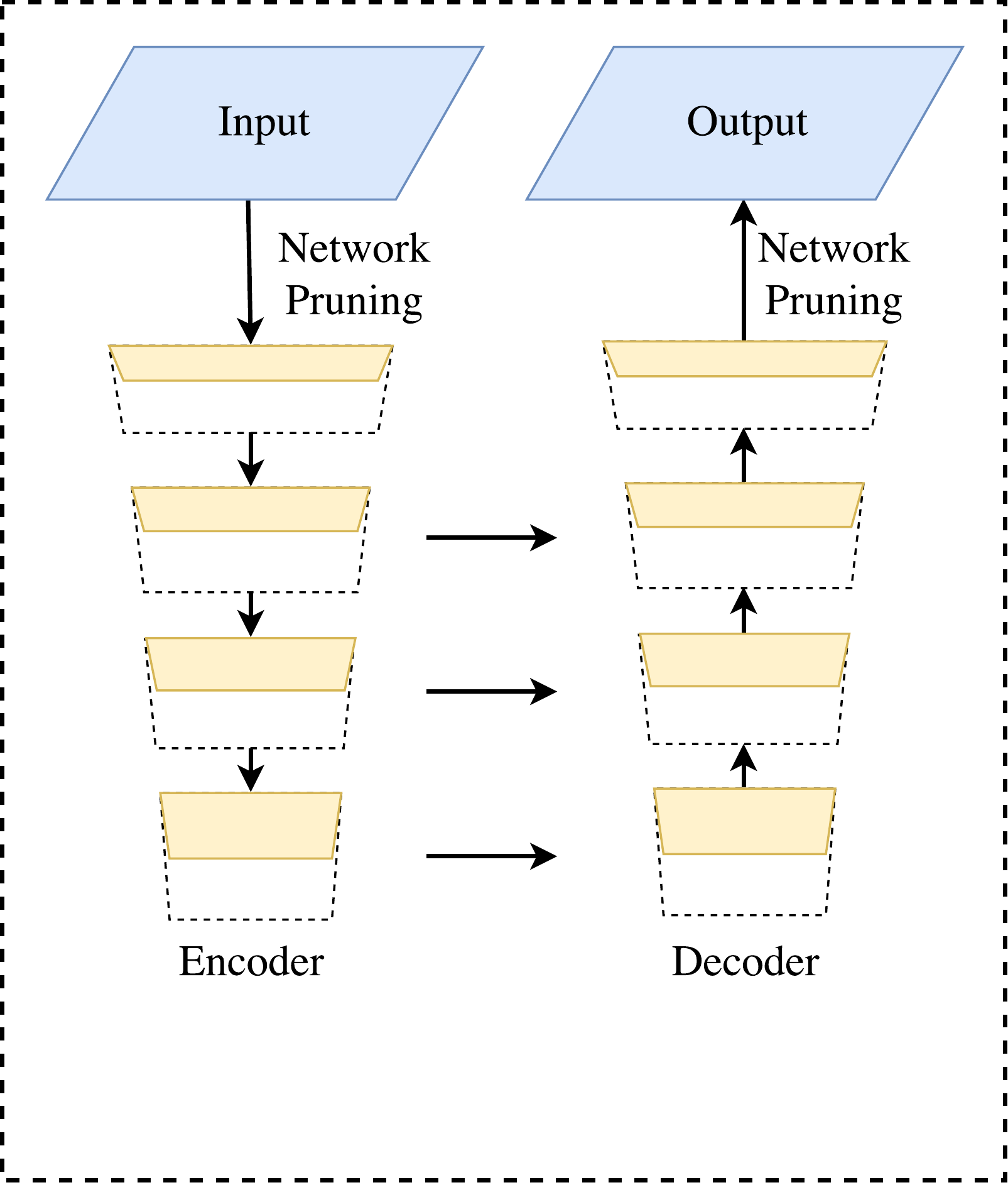}
			\end{minipage}
		}
		\hspace{-2ex}
		\subfigure[Ours]
		{
			\begin{minipage}{0.315\textwidth}
				\label{fig:dual-path}
				\centering
				\includegraphics[width=1.0\linewidth]{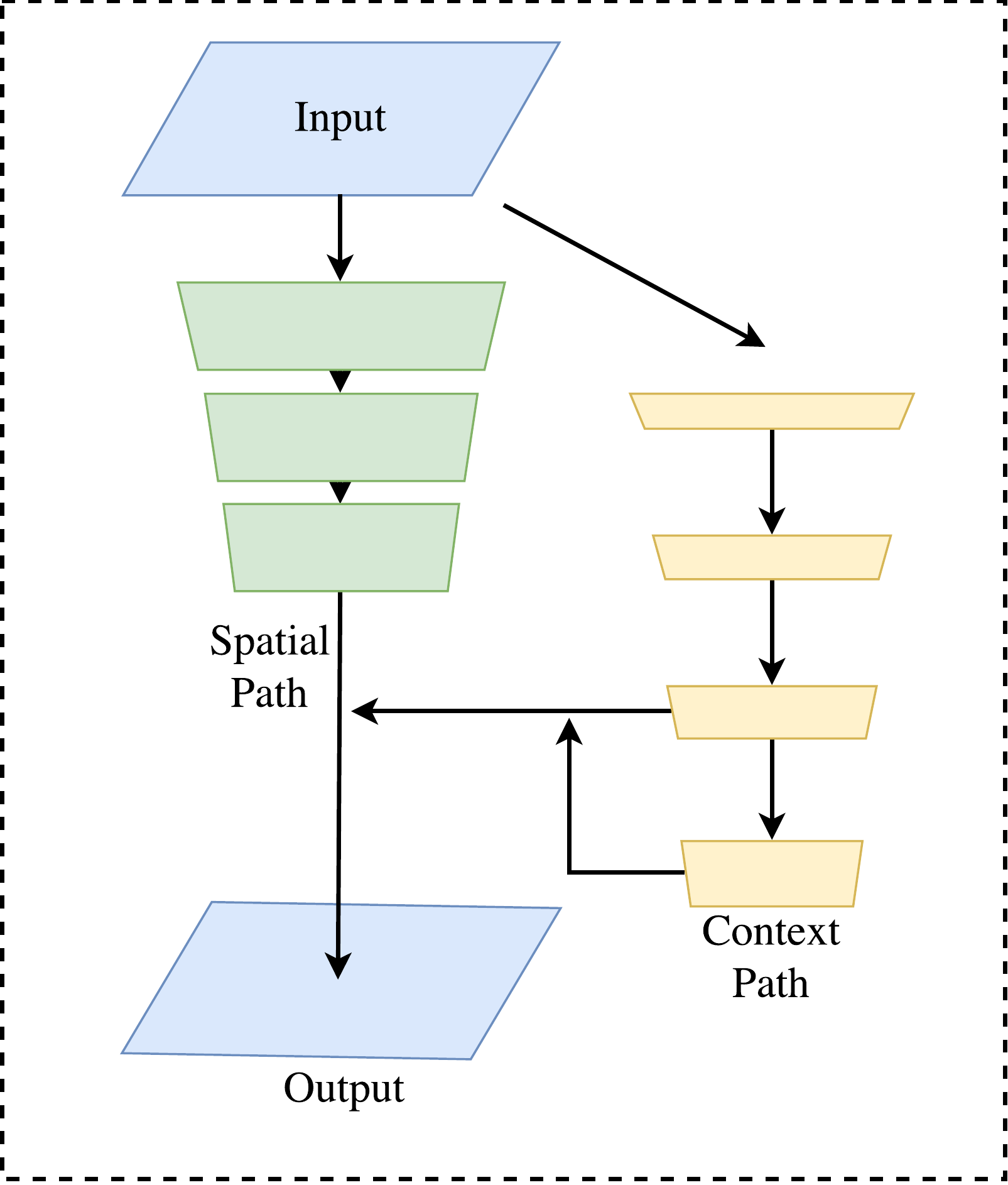}
			\end{minipage}
		}
		\caption{Illustration of the architectures to speed up and our proposed approach. (a) presents the cropping or resizing operation on the input image and the lightweight model with pruning channels or dropping stages. (b) indicates the U-shape structure. (c) demonstrates our proposed Bilateral Segmentation Network~(BiSeNet). The black dash line represents the operations which damage the spatial information, while the red dash line represents the operations which shrink the receptive field. The green block is our proposed Spatial Path~(SP). In the network part, each block represents the feature map of different down-sampling size. And the length of the block represents the spatial resolution, while the thickness is on behalf of the number of channels. }
	\end{figure}
	3) For the last case, ENet~\cite{Paszke-Arxiv-ENet-2016} proposes to drop the last stage of the model in pursuit of an extremely tight framework. Nevertheless, the drawback of this method is obvious: since the ENet abandons the downsampling operations in the last stage, the receptive field of the model is not enough to cover large objects, resulting in a poor discriminative ability. Overall, all of the above methods compromise the accuracy to speed, which is inferior in practice. Figure~\ref{fig:input-model} gives the illustration.
	
	To remedy the loss of spatial details mentioned above, researchers widely utilize the U-shape structure~\cite{Badrinarayanan-PAMI-SegNet-2017, Paszke-Arxiv-ENet-2016, Xie-ICCV-HEAD-2015}. By fusing the hierarchical features of the backbone network, the U-shape structure gradually increases the spatial resolution and fills some missing details. However, this technique has two weaknesses. 1) The complete U-shape structure can reduce the speed of the model due to the introduction of extra computation on high-resolution feature maps. 2) More importantly, most spatial information lost in the pruning or cropping cannot be easily recovered by involving the shallow layers as shown in Figure~\ref{fig:u-shape}. In other words, the U-shape technique is better to regard as a relief, rather than an essential solution.
	
	Based on the above observation, we propose the Bilateral Segmentation Network (BiSeNet) with two parts: \emph{Spatial Path} (SP) and \emph{Context Path} (CP). As their names imply, the two components are devised to confront with the loss of spatial information and shrinkage of receptive field respectively. The design philosophy of the two paths is clear. For \emph{Spatial Path}, we stack only three convolution layers to obtain the $1/8$ feature map, which retains affluent spatial details. In respect of \emph{Context Path}, we append a global average pooling layer on the tail of Xception~\cite{Chollet-CVPR-Xception-2017}, where the receptive field is the maximum of the backbone network. Figure~\ref{fig:dual-path} shows the structure of these two components.
	
	In pursuit of better accuracy without loss of speed, we also research the fusion of two paths and refinement of final prediction and propose Feature Fusion Module~(FFM) and Attention Refinement Module~(ARM) respectively. As our following experiments show, these two extra components can further improve the overall semantic segmentation accuracy on both Cityscapes~\cite{Cityscapes}, CamVid~\cite{CamVid}, and COCO-Stuff~\cite{Caesar-Stuff-2018} benchmarks.
	
	Our main contributions are summarized as follows:
	\begin{compactitem}
		\item We propose a novel approach to decouple the function of spatial information preservation and receptive field offering into two paths. Specifically, we propose a Bilateral Segmentation Network~(BiSeNet) with a Spatial Path~(SP) and a Context Path~(CP).
		\item We design two specific modules, Feature Fusion Module~(FFM) and Attention Refinement Module~(ARM), to further improve the accuracy with acceptable cost.
		\item We achieve impressive results on the benchmarks of Cityscapes, CamVid, and COCO-Stuff. More specifically, we obtain the results of 68.4\% on the Cityscapes test dataset with the speed of 105 FPS.
	\end{compactitem}
	
	\section{Related Work}
	\label{sec:related-work}
	Recently, lots of approaches based on FCN~\cite{Long-CVPR-FCN-2015} have achieved the \emph{state-of-the-art} performance on different benchmarks of the semantic segmentation task. Most of these methods are designed to encode more spatial information or enlarge the receptive field.
	
	\textbf{Spatial information:}
	The convolutional neural network~(CNN)~\cite{Krizhevsky-NIPS-Imagenet} encodes high-level semantic information with consecutive down-sampling operations. However, in the semantic segmentation task, the spatial information of the image is crucial to predicting the detailed output. Modern existing approaches devote to encode affluent spatial information. DUC~\cite{Wang-CVPR-DUC-2017}, PSPNet~\cite{Zhao-CVPR-PSPNet-2017}, DeepLab~v2~\cite{Chen-Arxiv-Deeplabv2-2016}, and Deeplab~v3~\cite{Chen-Arxiv-Deeplabv3-2017} use the dilated convolution to preserve the spatial size of the feature map. Global Convolution Network~\cite{Peng-CVPR-Largekernl-2017} utilizes the ``large kernel'' to enlarge the receptive field. 
	
	\textbf{U-Shape method:} 
	The U-shape structure~\cite{Badrinarayanan-PAMI-SegNet-2017,Noh-ICCV-Deconv-2015, Ronneberger-ICCV-U-net-2015, Long-CVPR-FCN-2015, Ghiasi-ECCV-LRR-2016} can recover a certain extent of spatial information. The original FCN~\cite{Long-CVPR-FCN-2015} network encodes different level features by a skip-connected network structure. Some methods employ their specific refinement structure into U-shape network structure. \cite{Badrinarayanan-PAMI-SegNet-2017,Noh-ICCV-Deconv-2015} create a U-shape network structure with the usage of deconvolution layers. U-net~\cite{Ronneberger-ICCV-U-net-2015} introduces the useful skip connection network structure for this task. Global Convolution Network~\cite{Peng-CVPR-Largekernl-2017} combines the U-shape structure with ``large kernel''. LRR~\cite{Ghiasi-ECCV-LRR-2016} adopts the Laplacian Pyramid Reconstruction Network. RefineNet~\cite{Lin-CVPR-Refinenet-2017} adds multi-path refinement structure to refine the prediction. DFN~\cite{Yu-CVPR-DFN-2018} designs a channel attention block to achieve the feature selection. However, in the U-shape structure, some lost spatial information cannot be easily recovered.
	
	\textbf{Context information:} Semantic segmentation requires context information to generate a high-quality result. The majority of common methods enlarge the receptive field or fuse different context information. \cite{Chen-Arxiv-Deeplabv2-2016, Chen-Arxiv-Deeplabv3-2017, Yu-ICLR-Dilate-2016, Wang-CVPR-DUC-2017} employ the different dilation rates in convolution layers to capture diverse context information. Driven by the image pyramid, multi-scale feature ensemble is always employed in the semantic segmentation network structure.  In~\cite{Chen-Arxiv-Deeplabv2-2016}, an ``ASPP'' module is proposed to capture context information of different receptive field. PSPNet~\cite{Zhao-CVPR-PSPNet-2017} applies a ``PSP'' module which contains several different scales of average pooling layers. ~\cite{Chen-Arxiv-Deeplabv3-2017} designs an ``ASPP'' module with global average pooling to capture the global context of the image. \cite{Zhang-ICCV-SAC-2017} improves the neural network by a scale adaptive convolution layer to obtain an adaptive field context information. DFN~\cite{Yu-CVPR-DFN-2018} adds the global pooling on the top of the U-shape structure to encode the global context.
	
	\textbf{Attention mechanism:}
	Attention mechanism can use the high-level information to guide the feed-forward network~\cite{Wang-CVPR-ResAttention-2017,Mnih-NIPS-RecurrentAttention-2014}. In~\cite{Chen-CVPR-AttentionScale-2016}, the attention of CNN depends on the scale of the input image. In~\cite{Hu-Arxiv-SEnet-2017}, they apply channel attention to recognition task and achieve the \emph{state-of-the-art}. Like the DFN~\cite{Yu-CVPR-DFN-2018}, they learn the global context as attention and revise the features.
	
	\textbf{Real time segmentation:}
	Real-time semantic segmentation algorithms require a fast way to generate the high-quality prediction. SegNet~\cite{Badrinarayanan-PAMI-SegNet-2017} utilizes a small network structure and the skip-connected method to achieve a fast speed. E-Net~\cite{Paszke-Arxiv-ENet-2016} designs a lightweight network from scratch and delivers an extremely high speed. ICNet~\cite{Zhao-Arxiv-ICNet-2017} uses the image cascade to speed up the semantic segmentation method. ~\cite{Li-CVPR-DLC-2017} employs a cascade network structure to reduce the computation in ``easy regions''. ~\cite{Wu-Arxiv-Sparsity-2017} designs a novel two-column network and spatial sparsity to reduce computation cost. Differently, our proposed method employs a lightweight model to provide sufficient receptive field. Furthermore, we set a shallow but wide network to capture adequate spatial information.
	
	\section{Bilateral Segmentation Network}
	\label{sec:method}
	In this section, we first illustrate our proposed Bilateral Segmentation Network~(BiSeNet) with Spatial Path and Context Path in detail. Furthermore, we elaborate on the effectiveness of these two paths correspondingly. Finally, we demonstrate how to combine the features of these two paths with Feature Fusion Module and the whole architecture of our BiSeNet.
	
	\subsection{Spatial path}
	\label{sec:spatial-path}
	In the task of semantic segmentation, some existing approaches~\cite{Wang-CVPR-DUC-2017, Chen-Arxiv-Deeplabv2-2016, Chen-Arxiv-Deeplabv3-2017, Zhao-CVPR-PSPNet-2017} attempt to preserve the resolution of the input image to encode enough spatial information with dilated convolution, while a few approaches~\cite{Chen-Arxiv-Deeplabv2-2016, Chen-Arxiv-Deeplabv3-2017, Zhao-CVPR-PSPNet-2017, Peng-CVPR-Largekernl-2017} try to capture sufficient receptive field with pyramid pooling module, atrous spatial pyramid pooling or ``large kernel''. These methods indicate that the spatial information and the receptive field are crucial to achieving high accuracy. However, it is hard to meet these two demands simultaneously. Especially, in the case of real-time semantic segmentation, existing modern approaches~\cite{Badrinarayanan-PAMI-SegNet-2017, Paszke-Arxiv-ENet-2016, Zhao-Arxiv-ICNet-2017} utilize small input image or lightweight base model to speed up. The small size of the input image loses the majority of spatial information from the original image, while the lightweight model damages spatial information with the channel pruning.
	
\begin{figure}[t]
		\centering
		\includegraphics[width=0.9\linewidth]{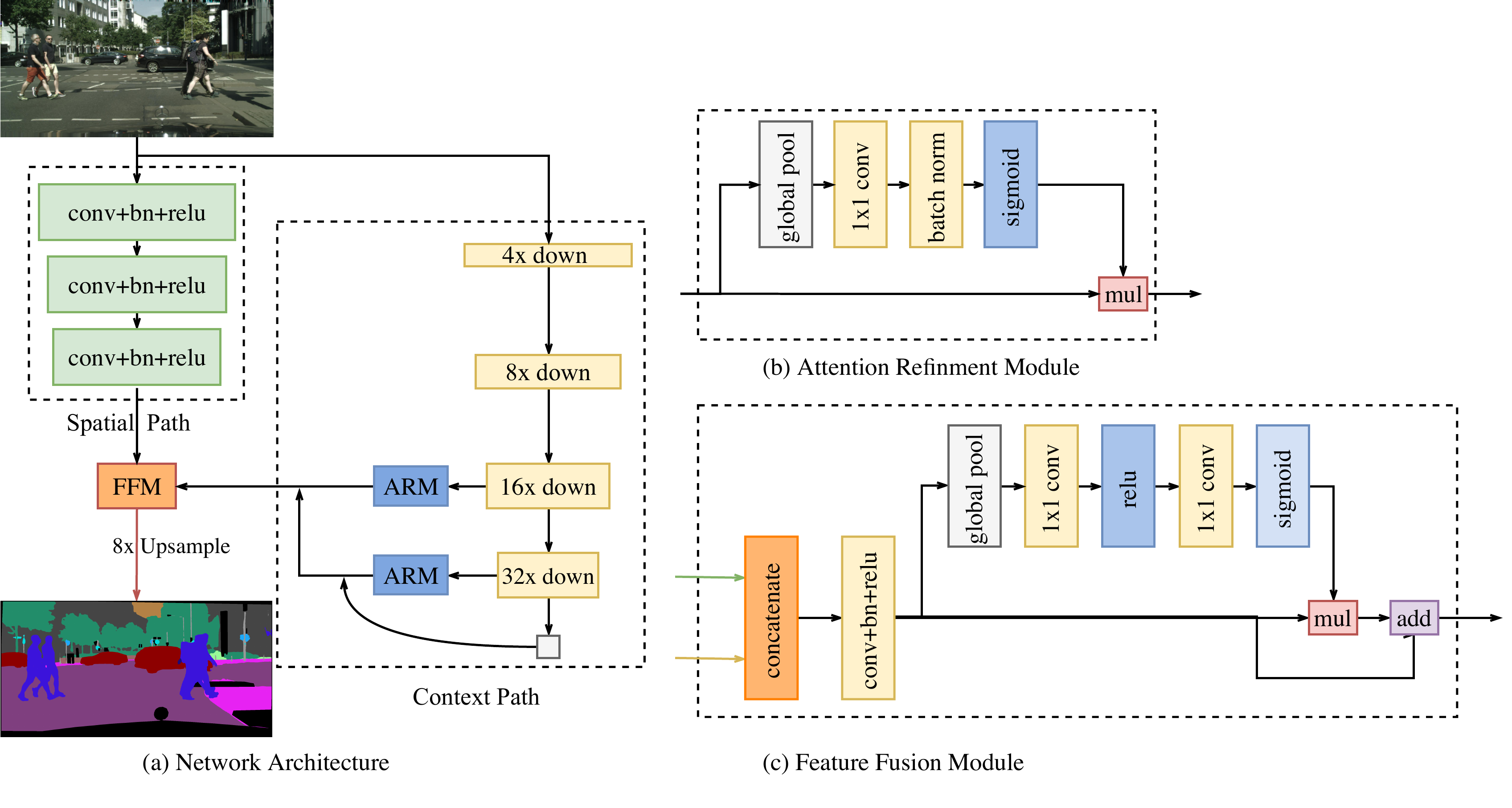}
		\caption{An overview of the Bilateral Segmentation Network. 
			(a) Network Architecture. The length of block indicates the spatial size, while the thickness represents the number of channels. 
			(b) Components of the Attention Refinement Module~(ARM).
			(c) Components of the Feature Fusion Module~(FFM). The read line represents we take this process only when testing.}
		\label{fig:network}
		\centering
	\end{figure}

	Based on this observation, we propose a Spatial Path to preserve the spatial size of the original input image and encode affluent spatial information. The Spatial Path contains three layers. Each layer includes a convolution with $stride=2$, followed by batch normalization~\cite{Ioffe-ICML-BN-2015} and ReLU~\cite{Glorot-AISTATS-ReLU-2011}. Therefore, this path extracts the output feature maps that is 1/8 of the original image. It encodes rich spatial information due to the large spatial size of feature maps. Figure~\ref{fig:network}(a) presents the details of the structure.
	
	\subsection{Context path}
	\label{sec:context-path}
	While the Spatial Path encodes affluent spatial information, the Context Path is designed to provide sufficient receptive field. In the semantic segmentation task, the receptive field is of great significance for the performance. To enlarge receptive field, some approaches have taken advantage of the pyramid pooling module~\cite{Zhao-CVPR-PSPNet-2017}, atrous spatial pyramid pooling~\cite{Chen-Arxiv-Deeplabv2-2016, Chen-Arxiv-Deeplabv3-2017} or ``large kernel''~\cite{Peng-CVPR-Largekernl-2017}. However, these operations are computation demanding and memory consuming, which result in the low speed. 
	
	With the consideration of the large receptive field and efficient computation simultaneously, we propose the Context Path. The Context Path utilizes lightweight model and global average pooling~\cite{Liu-ICLR-ParseNet-2016, Chen-Arxiv-Deeplabv2-2016, Chen-Arxiv-Deeplabv3-2017} to provide large receptive field. In this work, the lightweight model, like Xception~\cite{Chollet-CVPR-Xception-2017}, can downsample the feature map fast to obtain large receptive field, which encodes high level semantic context information. Then we add a global average pooling on the tail of the lightweight model, which can provide the maximum receptive field with global context information. Finally, we combine the up-sampled output feature of global pooling and the features of the lightweight model. In the lightweight model, we deploy U-shape structure~\cite{Xie-ICCV-HEAD-2015, Badrinarayanan-PAMI-SegNet-2017, Paszke-Arxiv-ENet-2016} to fuse the features of the last two stages, which is an incomplete U-shape style. Figure~\ref{fig:network}(c) shows the overall perspective of the Context Path.
	
	\paragraph{Attention refinement module:}
	In the Context Path, we propose a specific Attention Refinement Module~(ARM) to refine the features of each stage. As Figure~\ref{fig:network}(b) shows, ARM employs global average pooling to capture global context and computes an attention vector to guide the feature learning. This design can refine the output feature of each stage in the Context Path. It integrates the global context information easily without any up-sampling operation. Therefore, it demands negligible computation cost.
	
	\subsection{Network architecture}
	\label{sec:network}
	With the Spatial Path and the Context Path, we propose BiSeNet for real-time semantic segmentation as illustrated in Figure~\ref{fig:network}(a).  
	
	We use the pre-trained Xception model as the backbone of the Context Path and three convolution layers with stride as the Spatial Path. And then we fuse the output features of these two paths to make the final prediction. It can achieve real-time performance and high accuracy at the same time. First, we focus on the practical computation aspect. Although the Spatial Path has large spatial size, it only has three convolution layers. Therefore, it is not computation intensive. As for the Context Path, we use a lightweight model to down-sample rapidly. Furthermore, these two paths compute concurrently, which considerably increase the efficiency. Second, we discuss the accuracy aspect of this network. In our paper, the Spatial Path encodes rich spatial information, while the Context Path provides large receptive field. They are complementary to each other for higher performance.
	
	\paragraph{Feature fusion module:}
	The features of the two paths are different in level of feature representation. Therefore, we can not simply sum up these features. The spatial information captured by the Spatial Path encodes mostly rich detail information. Moreover, the output feature of the Context Path mainly encodes context information. In other words, the output feature of Spatial Path is low level, while the output feature of Context Path is high level. Therefore, we propose a specific Feature Fusion Module to fuse these features.
	
	Given the different level of the features, we first concatenate the output features of Spatial Path and Context Path. And then we utilize the batch normalization~\cite{Ioffe-ICML-BN-2015} to balance the scales of the features. Next, we pool the concatenated feature to a feature vector and compute a weight vector, like SENet~\cite{Hu-Arxiv-SEnet-2017}. This weight vector can re-weight the features, which amounts to feature selection and combination. Figure~\ref{fig:network}(c) shows the details of this design.
	
	\paragraph{Loss function:}
	In this paper, we also utilize the auxiliary loss function to supervise the training of our proposed method. We use the principal loss function to supervise the output of the whole BiSeNet. Moreover, we add two specific auxiliary loss functions to supervise the output of the Context Path, like deep supervision~\cite{Xie-ICCV-HEAD-2015}. All the loss functions are Softmax loss, as Equation~\ref{eqn:softmax} shows. Furthermore, we use the parameter $\alpha$ to balance the weight of the principal loss and auxiliary loss, as Equation~\ref{eqn:combination} presents. The $\alpha$ in our paper is equal to 1. The joint loss makes optimizer more comfortable to optimize the model.
	
	\begin{equation}
	\label{eqn:softmax}
	loss = \frac{1}{N}\sum_{i}{L_i}=\frac{1}{N}\sum_{i}{-log\left(\frac{e^{p_i}}{\sum_j{e^{p_j}}}\right)}
	\end{equation}
	where $p$ is the output prediction of the network.
	
	\begin{equation}
	\label{eqn:combination}
	L(X;W) = l_{p}(X;W) + \alpha \sum_{i=2}^{K}l_i(X_i;W)
	\end{equation}
	where $l_{p}$ is the principal loss of the concatenated output. $X_i$ is the output feature from stage $i$ of Xception model. $l_{i}$ is the auxiliary loss for stage $i$. The $K$ is equal to 3 in our paper. The $L$ is the joint loss function. Here, we only use the auxiliary loss in the training phase.

	\section{Experimental Results}
	\label{sec:exp}
	We adopt a modified Xception model~\cite{Chollet-CVPR-Xception-2017}, Xception39, into the real-time semantic segmentation task. Our implementation code will be made publicly available. 
	
	We evaluate our proposed BiSeNet on Cityscapes~\cite{Cityscapes}, CamVid~\cite{CamVid} and COCO-Stuff~\cite{Caesar-Stuff-2018} benchmarks. We first introduce the datasets and the implementation protocol. Next, we describe our speed strategy in comparison with other methods in detail. And then we investigate the effects of each component of our proposed approach. We evaluate all performance results on the Cityscapes validation set. Finally, we report the accuracy and speed results on Cityscapes, CamVid and COCO-Stuff datasets compared with other real-time semantic segmentation algorithms.
	
	\paragraph{Cityscapes:}
	The Cityscapes~\cite{Cityscapes} is a large urban street scene dataset from a car perspective. It contains 2,975 fine annotated images for training and another 500 images for validation. In our experiments, we only use the fine annotated images. For testing, it offers 1,525 images without ground-truth for fair comparison. These images all have a resolution of 2,048$\times$1,024, in which each pixel is annotated to pre-defined 19 classes.


	\paragraph{CamVid:}
	The CamVid~\cite{CamVid} is another street scene dataset from the perspective of a driving automobile. It contains 701 images in total, in which 367 for training, 101 for validation and 233 for testing. The images have a resolution of 960$\times$720 and 11 semantic categories.
	
	\paragraph{COCO-Stuff:}
    The COCO-Stuff~\cite{Caesar-Stuff-2018} augments all 164,000 images of the popular COCO~\cite{Lin-COCO-2014} dataset, out of which 118,000 images for training, 5,000 images for validation, 20,000 images for test-dev and 20,000 images for test-challenge. It covers 91 stuff classes and 1 class 'unlabeled'. 
	
	\subsection{Implementation protocol}
	In this section, we elaborate our implementation protocol in detail. 
	
	\paragraph{Network:}
	We apply three convolutions as Spatial Path and Xception39 model for Context Path. And then we use Feature Fusion Module to combine the features of these two paths to predict the final results. The output resolution of Spatial Path and the final prediction are 1/8 of the original image.
	
	\paragraph{Training details:}
	We use mini-batch stochastic gradient descent~(SGD)~\cite{Krizhevsky-NIPS-Imagenet} with batch size 16, momentum $0.9$ and weight decay $1e^{-4}$ in training. Similar to ~\cite{Liu-ICLR-ParseNet-2016, Chen-Arxiv-Deeplabv2-2016, Chen-Arxiv-Deeplabv3-2017}, we apply the ``poly'' learning rate strategy in which the initial rate is multiplied by $(1 - \frac{iter}{max\_iter})^{power}$ each iteration with power $0.9$. The initial learning rate is $2.5e^{-2}$.  
	
	
	\paragraph{Data augmentation:} 
	We employ the mean subtraction, random horizontal flip and random scale on the input images to augment the dataset in training process. The scales contains \{ 0.75, 1.0, 1.5, 1.75, 2.0\}. Finally, we randomly crop the image into fix size for training.
	
	
	\subsection{Ablation study}
	In this subsection, we detailedly investigate the effect of each component in our proposed BiSeNet step by step. In the following experiments, we use Xception39 as the base network and evaluate our method on the Cityscapes validation dataset~\cite{Cityscapes}.
	
	\setlength{\tabcolsep}{4pt}
	\begin{table}[!t]
		\begin{center}
			\caption{Accuracy and parameter analysis of our baseline model: Xception39 and Res18 on Cityscapes validation dataset. Here we use FCN-32s as the base structure. FLOPS are estimated for input of $3\times640\times360$.}
			\label{tab:performance-baseline}
			\begin{tabular}{lcccc}
				\toprule
				Method & BaseModel & FLOPS & Parameters & Mean IOU(\%)\\
				\noalign{\smallskip}
				\midrule
				\noalign{\smallskip}
				FCN-32s & Xception39 & 185.5M & 1.2M & 60.78 \\
				FCN-32s & Res18 & 8.3G & 42.7M & 61.58\\
				\bottomrule
			\end{tabular}
		\end{center}
	\end{table}
	\setlength{\tabcolsep}{1.4pt}
	
	\paragraph{Baseline:}
	We use the Xception39 network pretrained on Image{N}et dataset~\cite{ILSVRC15} as the backbone of Context Path. And then we directly up-sample the output of the network as original input image, like FCN~\cite{Long-CVPR-FCN-2015}. We evaluate the performance of the base model as our baseline, as shown in Table~\ref{tab:performance-baseline}.
	
	\setlength{\tabcolsep}{4pt}
	\begin{table}[!t]
		\begin{center}
			\caption{Speed analysis of the U-shape-8s and the U-shape-4s on one NVIDIA Titan XP card. Image size is W$\times$H.}
			\label{tab:ushape-speed}
			\begin{tabular}{cccccccccccccccc}
				\toprule
				\multicolumn{2}{c}{\multirow{3}*{Method}} & \multicolumn{6}{c}{NVIDIA Titan XP} & \multicolumn{2}{c}{\multirow{3}*{Mean IOU(\%)}}\\
				\cmidrule(lr){3-8}
				\multicolumn{2}{c}{} & \multicolumn{2}{c}{640$\times$360} & \multicolumn{2}{c}{1280$\times$720} & \multicolumn{2}{c}{1920$\times$1080} & \\
				\multicolumn{2}{c}{} & ms & fps & ms & fps & ms & fps \\
				\noalign{\smallskip}
				\midrule
				\noalign{\smallskip}
				\multicolumn{2}{l}{U-shape-8s} & 3 & 413.7 & 6  &189.8 & 12 & 86.7 & \multicolumn{2}{c}{66.01} \\
				\multicolumn{2}{l}{U-shape-4s} & 4 &322.9 &  9 & 114& 17 &  61.1 & \multicolumn{2}{c}{66.13}\\
				\bottomrule
			\end{tabular}
		\end{center}
	\end{table}
	\setlength{\tabcolsep}{1.4pt}
	
	\paragraph{Ablation for U-shape:}
	We propose the Context Path to provide sufficient receptive field. where we use a lightweight model, Xception39, as the backbone of Context Path to down-sample quickly. Simultaneously, we use the U-shape structure~\cite{Xie-ICCV-HEAD-2015, Badrinarayanan-PAMI-SegNet-2017, Paszke-Arxiv-ENet-2016} to combine the features of the last two stage in Xception39 network, called U-shape-8s, rather than the standard U-shape structure, called U-shape-4s. The number represents the down-sampling factor of the output feature, as shown in Figure~\ref{fig:network}. The reason to use U-shape-8s structure is twofold. First, the U-shape structure can recover a certain extent of spatial information and spatial size. Second, the U-shape-8s structure is faster compared to the U-shape-4s, as shown in Table~\ref{tab:ushape-speed}. Therefore, we use the U-shape-8s structure, which improves the performance from $60.79\%$ to $66.01\%$, as shown in Table~\ref{tab:ushape-speed}. 
	
	\begin{figure}
		\centering
		\vspace{.3in}
		\subfigure[Image]
		{
			\begin{minipage}{0.22\textwidth}
				\centering
				\includegraphics[width=1.0\linewidth]{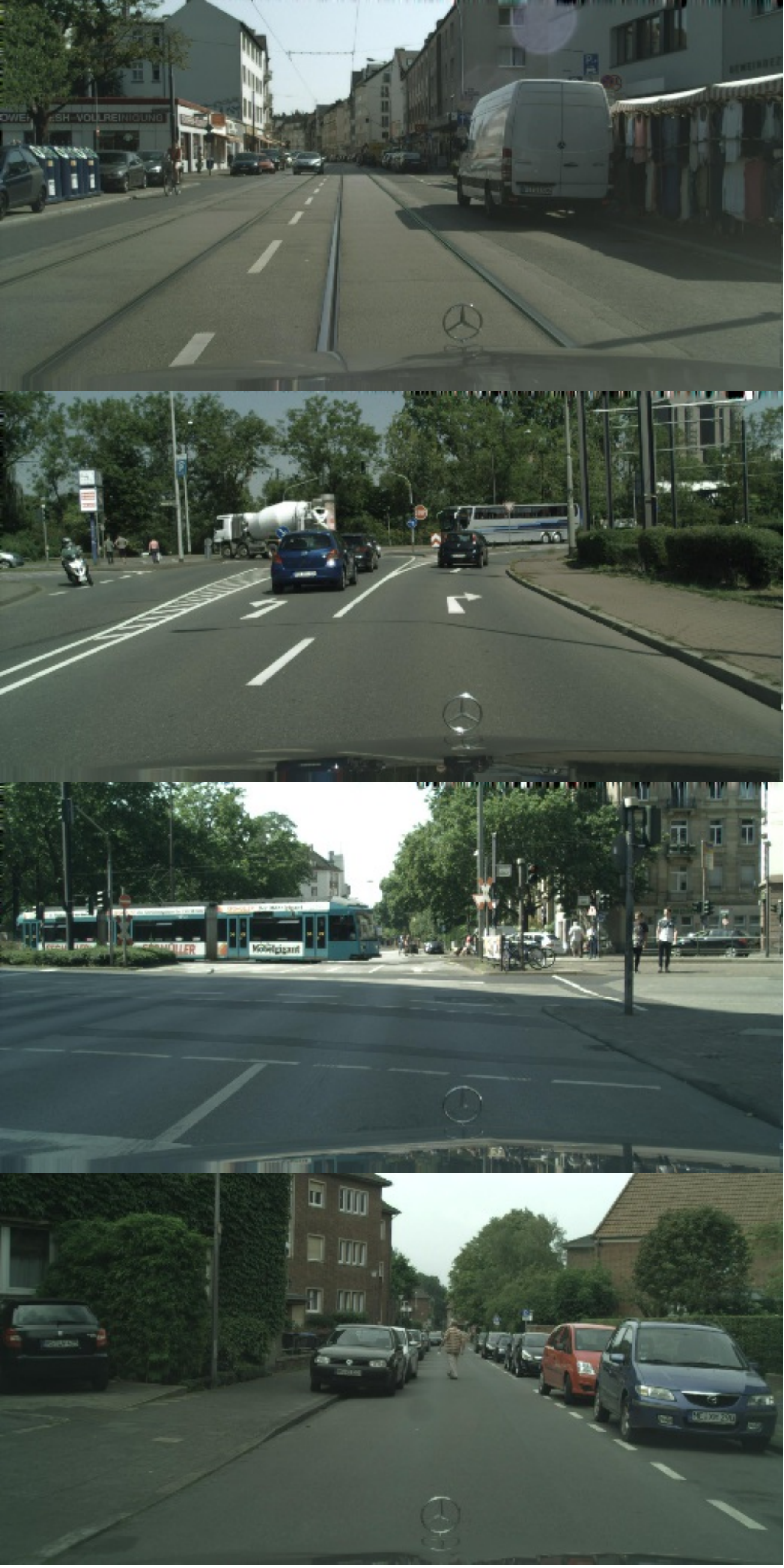}
				\label{fig:mid_img}
			\end{minipage}
		}
		\hspace{-3ex}
		\subfigure[U-Shape]
		{
			\begin{minipage}{0.22\textwidth}
				\centering
				\includegraphics[width=1.0\linewidth]{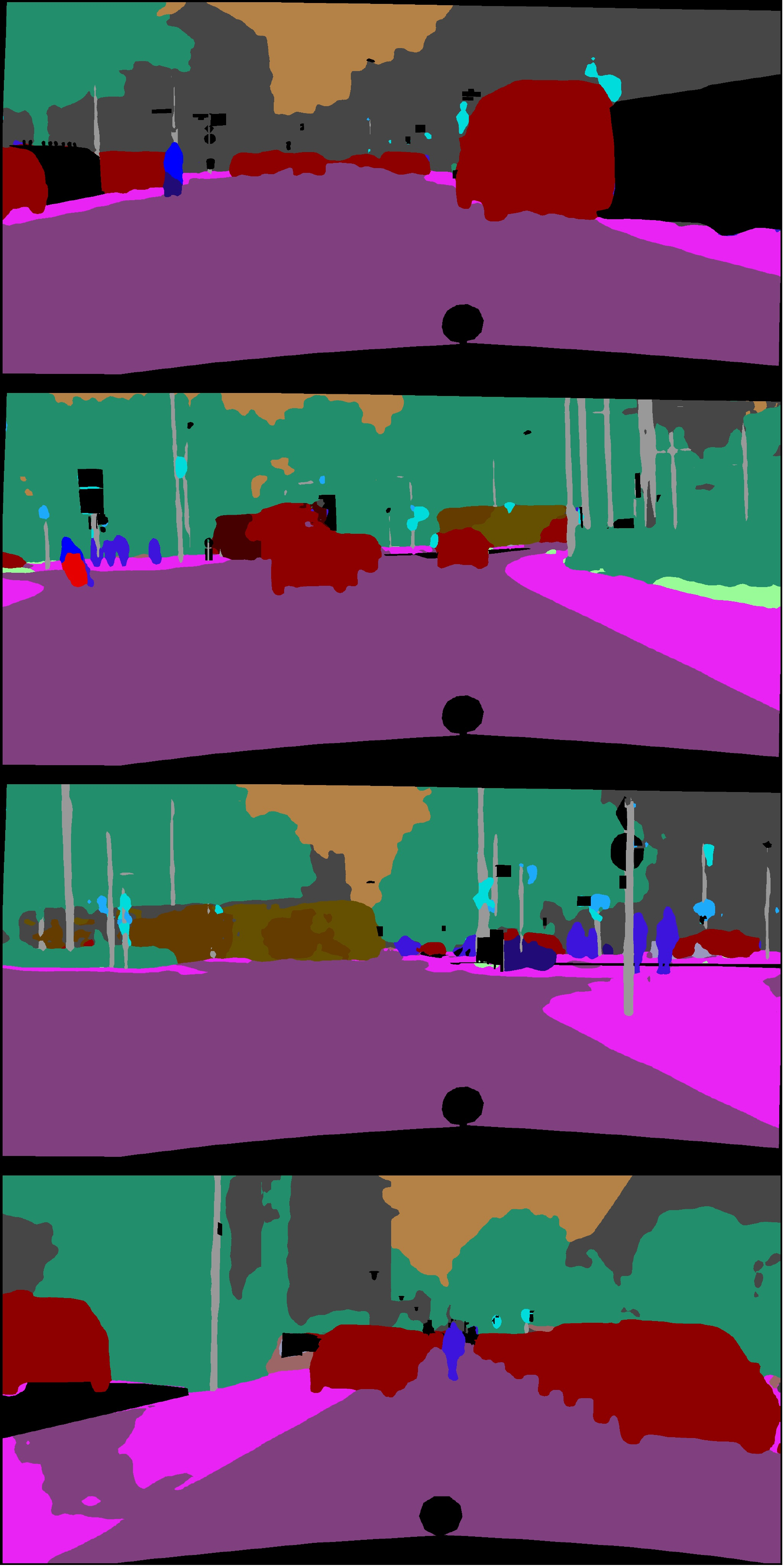}
				\label{fig:mid_ushape}
			\end{minipage}
		}
		\hspace{-3ex}
		\subfigure[BiSeNet]
		{
			\begin{minipage}{0.22\textwidth}
				\centering
				\includegraphics[width=1.0\linewidth]{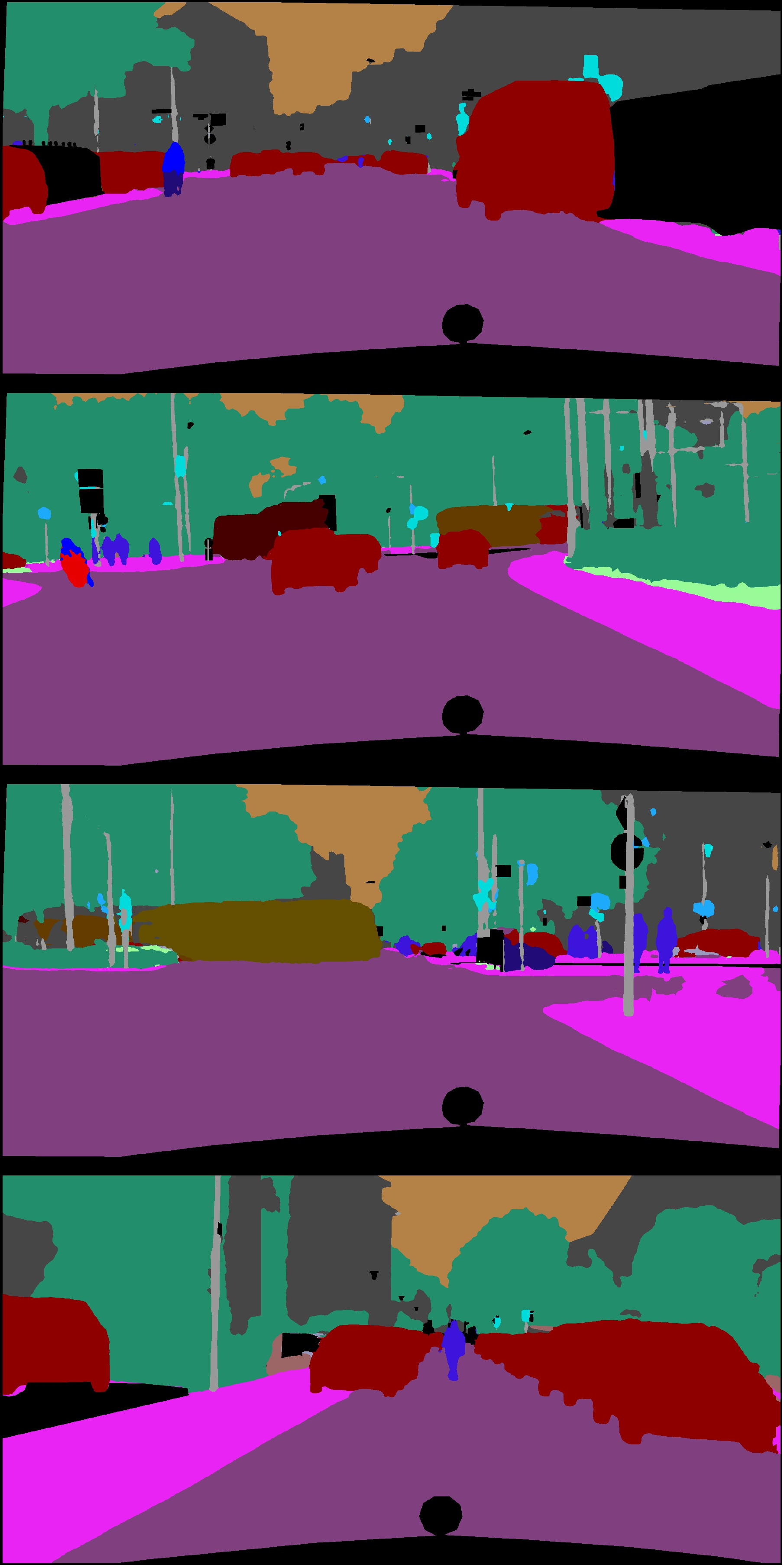}
				\label{fig:mid_dualpath}
			\end{minipage}
		}
		\hspace{-3ex}
		\subfigure[GT]
		{
			\begin{minipage}{0.22\textwidth}
				\centering
				\includegraphics[width=1.0\linewidth]{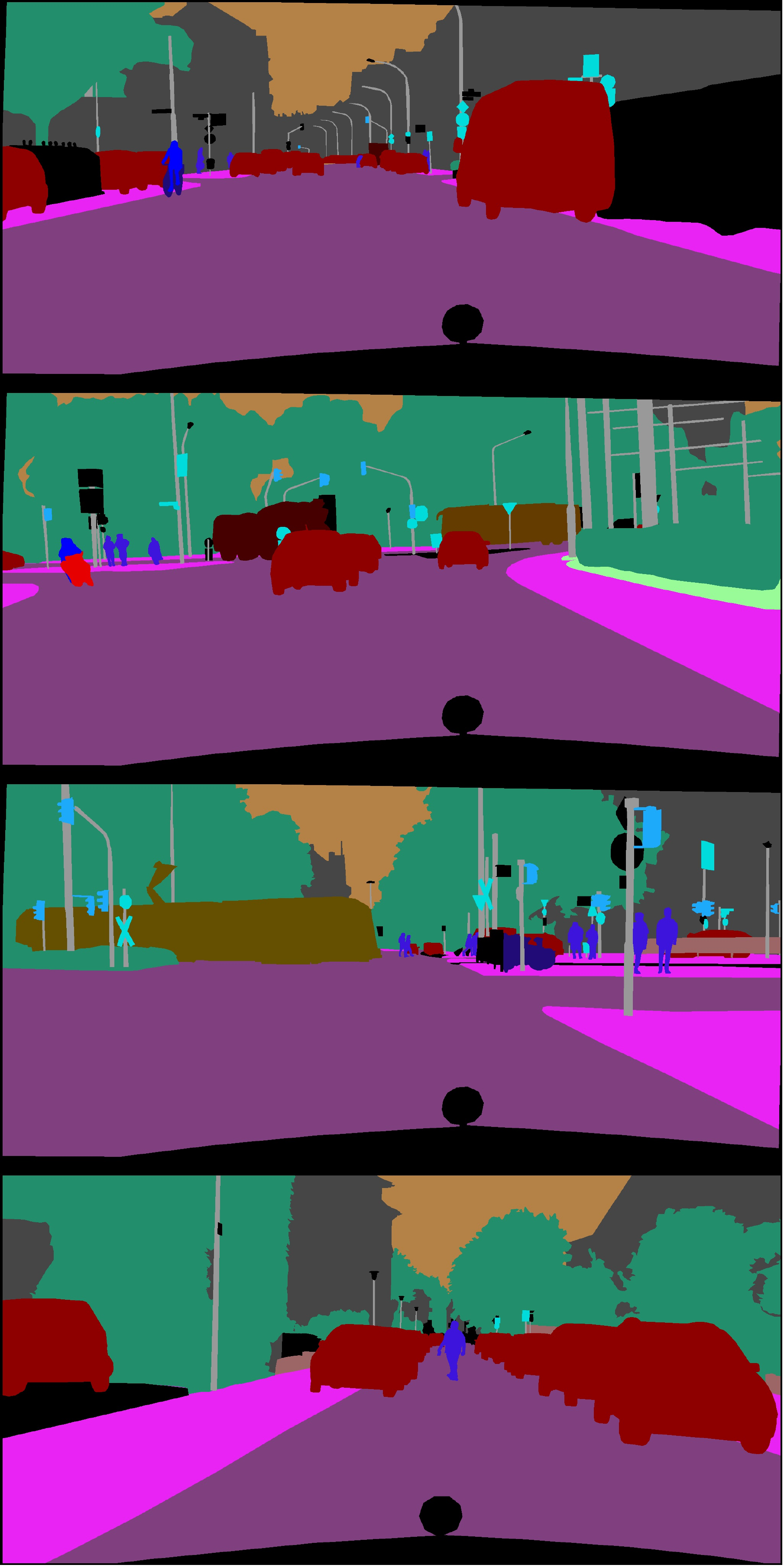}
				\label{fig:mid_gt}
			\end{minipage}
		}
		\caption{Example results of the output before adding the Spatial Path and after adding the Spatial Path. The output BiSeNet has more detail information than the output of U-shape.}
		\label{fig:spatial_path}
	\end{figure}
	
	\paragraph{Ablation for spatial path:}
	As Section~\ref{sec:intro} stated, existing modern approaches of real-time semantic segmentation task face the challenge of lost of spatial information. Therefore, we propose a Spatial Path to preserve the spatial size and capture rich spatial information. The Spatial Path contains three convolutions with $stride=2$, followed by batch normalization~\cite{Ioffe-ICML-BN-2015} and ReLU~\cite{Glorot-AISTATS-ReLU-2011}. This improves the performance from $66.01\%$ to $67.42\%$, as shown in Table~\ref{tab:component}. The Spatial Path encodes abundant details of spatial information. Figure~\ref{fig:spatial_path} shows that the BiSeNet can obtain more detailed spatial information, e.g. some traffic signs.
	
	\setlength{\tabcolsep}{4pt}
	\begin{table}[!t]
		\begin{center}
			\caption{Detailed performance comparison of each component in our proposed BiSeNet. \textbf{CP}: Context Path; \textbf{SP}: Spatial Path; \textbf{GP}: global average pooling; \textbf{ARM}: Attention Refinement Module; \textbf{FFM}: Feature Fusion Module.}
			\label{tab:component}
			\begin{tabular}{lc}
				\toprule
				Method & Mean IOU(\%)\\
				\noalign{\smallskip}
				\midrule
				\noalign{\smallskip}
				CP & 66.01 \\
				CP+SP(Sum) & 66.82 \\
				CP+SP(FFM) & 67.42 \\
				\noalign{\smallskip}
				\midrule
				\noalign{\smallskip}
				CP+SP(FFM)+GP & 68.42 \\
				CP+SP(FFM)+ARM &  68.72 \\
				\noalign{\smallskip}
				\midrule
				\noalign{\smallskip}
				CP+SP(FFM)+GP+ARM & 71.40\\
				\bottomrule
			\end{tabular}
		\end{center}
	\end{table}
	\setlength{\tabcolsep}{1.4pt}
	
	\paragraph{Ablation for attention refinement module:}
	For further improving the performance, we specially design an Attention Refinement Module~(ARM). This module contains a global average pooling to encode a ouput feature into a vector. Then we utilize a convolution, batch normalization~\cite{Ioffe-ICML-BN-2015} and ReLU unit~\cite{Glorot-AISTATS-ReLU-2011} to compute the attention vector. The original feature will be re-weighted by the attention vector. For the original feature, it is easy to capture the global context information without the complex up-sample operation. The effect of the ARM is presented in Table~\ref{tab:component}.  
	
	\paragraph{Ablation for feature fusion module:}
	Based on the Spatial Path and Context Path, we need to fuse the output features of these two paths. With the consideration of the different levels of the features, low level for the features of Spatial Path and high level for the Context Path, we propose the Feature Fusion Module to combine these features effectively. First, we evaluate the effect of a straightforward sum of these features and our proposed Feature Fusion Module, as shown in Table~\ref{tab:component}. The gap of the comparison performance explains the features of the two paths belong to different levels in turn.
	
	\paragraph{Ablation for global average pooling:}
	We expect the Context Path can provide sufficient receptive field. Although the original Xception39 model can cover the most region of input image theoretically, we still enlarge the receptive field further with global average pooling~\cite{Liu-ICLR-ParseNet-2016}. This can ensure the valid receptive field is large enough. In this paper, we add the global average pooling at the tail of the Xception39 model. Then, we up-sample the output of the global average pooling and sum up this feature with the output of the last stage in the Xception39 model, like DFN~\cite{Yu-CVPR-DFN-2018}. This  improves the performance from $67.42\%$ to $68.42\%$, which indicates the effect of this design, as shown in Table~\ref{tab:component}.
	
	\setlength{\tabcolsep}{4pt}
	\begin{table}[!t]
		\begin{center}
			\caption{Accuracy and parameter analysis of our baseline model: Xception39 and Res18 on Cityscapes validation dataset. Here we use FCN-32s as the base structure. FLOPS are estimated for input of $3\times640\times360$.}
			\label{tab:parameter-comparison}
			\begin{tabular}{lccc}
				\toprule
				Method & BaseModel & GFLOPS & Parameters \\
				\noalign{\smallskip}
				\hline
				\noalign{\smallskip}
				SegNet~\cite{Badrinarayanan-PAMI-SegNet-2017} & VGG16~\cite{Simonyan-ICLR-VGG-2015} & 286.0 & 29.5M \\
				ENet~\cite{Paszke-Arxiv-ENet-2016} & From scratch & 3.8 & 0.4M \\
				\noalign{\smallskip}
				\hline
				\noalign{\smallskip}
				Ours & Xception39 & 2.9 & 5.8M \\ 
				Ours & Res18 & 10.8 & 49.0M \\
				\bottomrule
			\end{tabular}
		\end{center}
	\end{table}
	\setlength{\tabcolsep}{1.4pt}
	
	\setlength{\tabcolsep}{3.7pt}
	\begin{table}[!t]
		\begin{center}
			\caption{Speed comparison of our method against other \emph{state-of-the-art} methods. Image size is W$\times$H. The \emph{Ours$^1$} and \emph{Ours$^2$} are the BiSeNet based on Xception39 and Res18 model.}
			\label{tab:speed-comp}
			\begin{tabular}{lccccccccccccc}
				\toprule
				\multicolumn{2}{l}{\multirow{3}*{Method}} & \multicolumn{6}{c}{NVIDIA Titan X}& \multicolumn{6}{c}{NVIDIA Titan XP}\\
				\cmidrule(lr){3-8} \cmidrule(lr){9-14}  
				\multicolumn{2}{c}{} & \multicolumn{2}{c}{640$\times$360} & \multicolumn{2}{c}{1280$\times$720} & \multicolumn{2}{c}{1920$\times$1080} & \multicolumn{2}{c}{640$\times$360} & \multicolumn{2}{c}{1280$\times$720} & \multicolumn{2}{c}{1920$\times$1080} \\
				\multicolumn{2}{c}{} & ms & fps & ms & fps & ms & fps & ms & fps & ms & fps & ms & fps \\
				\noalign{\smallskip}
				\hline
				\noalign{\smallskip}
				\multicolumn{2}{l}{SegNet~\cite{Badrinarayanan-PAMI-SegNet-2017}} & 69 & 14.6 & 289 & 3.5 & 637 & 1.6 & - & - & - & - & - & - \\
				\multicolumn{2}{l}{ENet~\cite{Paszke-Arxiv-ENet-2016}} & 7 & 135.4 & 21 & 46.8 & 46 & 21.6 & - & - & - & - & - & - \\
				\noalign{\smallskip}
				\hline
				\noalign{\smallskip}
				\multicolumn{2}{l}{Ours$^1$} & \textbf{5} &\textbf{203.5} & \textbf{12} &\textbf{82.3} & \textbf{24} &\textbf{41.4} & \textbf{4} & \textbf{285.2} & \textbf{8} & \textbf{124.1} & \textbf{18} & \textbf{57.3} \\
				\multicolumn{2}{l}{Ours$^2$} & 8 & 129.4& 21 &47.9 & 43 &23 & 5 & 205.7 & 13 & 78.8 &29 & 34.4 \\
				\bottomrule
			\end{tabular}
		\end{center}
	\end{table}
	\setlength{\tabcolsep}{1.4pt}
	
	\subsection{Speed and Accuracy Analysis}
	In this section, we first analysis the speed of our algorithm. Then we report our final results on Cityscapes~\cite{Cityscapes}, CamVid~\cite{CamVid} and COCO-Stuff~\cite{Caesar-Stuff-2018} benchmarks compared with other algorithms.
	
	\paragraph{Speed analysis:}
	Speed is a vital factor of an algorithm especially when we apply it in practice. We conduct our experiments on different settings for thorough comparison. First, we show our status of FLOPS and parameters in Table~\ref{tab:parameter-comparison}. The FLOPS and parameters indicate the number of operations to process images of this resolution. For a fair comparison, we choose the 640$\times$360 as the resolution of the input image. Meanwhile, Table~\ref{tab:speed-comp} presents the speed comparison between our method with other approaches on different resolutions of input images and different hardware benchmarks. Finally, we report our speed and corresponding accuracy results on Cityscapes test dataset. From Table~\ref{tab:city-speed-comp}, we can find out our method achieves significant progress against the other methods both in speed and accuracy. In the evaluation process, we first scale the input image of 2048$\times$1024 resolution into the 1536$\times$768 resolution for testing the speed and accuracy. Meanwhile, we compute the loss function with the online bootstrapping strategy as described in \cite{Wu-Arxiv-HighPerf-2016}. In this process, we don't employ any testing technology, like multi-scale or multi-crop testing. 
	
	\setlength{\tabcolsep}{4pt}
	\begin{table}[!t]
		\begin{center}
			\caption{Accuracy and speed comparison of our method against other \emph{state-of-the-art} methods on Cityscapes test dataset. We train and evaluate on NVIDIA Titan XP with 2048$\times$1024 resolution input. ``-'' indicates that the methods didn't give the corresponding speed result of the accuracy.}
			\label{tab:city-speed-comp}
			\begin{tabular}{cccccccc}
				\toprule
				\multicolumn{2}{l}{\multirow{2}*{Method}} & \multicolumn{2}{l}{\multirow{2}*{BaseModel}} & \multicolumn{2}{c}{Mean IOU(\%)} & \multicolumn{2}{c}{\multirow{2}*{FPS}}\\
				\cmidrule(lr){5-6}
				\multicolumn{2}{c}{} & \multicolumn{2}{c}{} & \emph{val} & \emph{test} & \multicolumn{2}{c}{}\\
				\noalign{\smallskip}
				\hline
				\noalign{\smallskip}
				\multicolumn{2}{l}{SegNet~\cite{Badrinarayanan-PAMI-SegNet-2017}} & \multicolumn{2}{l}{VGG16} & - & 56.1 & \multicolumn{2}{c}{-}\\
				\multicolumn{2}{l}{ENet~\cite{Paszke-Arxiv-ENet-2016}} & \multicolumn{2}{l}{From scratch} & - &  58.3& \multicolumn{2}{c}{-}\\
				\multicolumn{2}{l}{SQ~\cite{Treml-NIPSW-SQ-2016}} & \multicolumn{2}{l}{SqueezeNet~\cite{Iandola-Arxiv-SqueezeNet-2016}} & - & 59.8 & \multicolumn{2}{c}{-}\\
				\multicolumn{2}{l}{ICNet~\cite{Zhao-Arxiv-ICNet-2017}} & \multicolumn{2}{l}{PSPNet50~\cite{Zhao-CVPR-PSPNet-2017}} & 67.7 & 69.5 & \multicolumn{2}{c}{30.3}\\
				\multicolumn{2}{l}{DLC~\cite{Li-CVPR-DLC-2017}} & \multicolumn{2}{l}{Inception-ResNet-v2} & - & 71.1 & \multicolumn{2}{c}{-}\\ 
				\multicolumn{2}{l}{Two-column Net~\cite{Wu-Arxiv-Sparsity-2017}} & \multicolumn{2}{l}{Res50} & \underline{74.6} & \underline{72.9}& \multicolumn{2}{c}{14.7}\\
				\noalign{\smallskip}
				\hline
				\noalign{\smallskip}
				\multicolumn{2}{l}{Ours} & \multicolumn{2}{l}{Xception39} & 69.0 &  68.4& \multicolumn{2}{c}{\textbf{105.8}}\\
				\multicolumn{2}{l}{Ours} & \multicolumn{2}{l}{Res18} &  \textbf{74.8} & \textbf{74.7} & \multicolumn{2}{c}{\underline{65.5}}\\
				\bottomrule
			\end{tabular}
		\end{center}
	\end{table}
	\setlength{\tabcolsep}{1.4pt}
	
		\begin{figure}[!t]
		\centering
		\subfigure[Image]
		{
			\begin{minipage}{0.18\textwidth}
				\centering
				\includegraphics[width=1.0\linewidth]{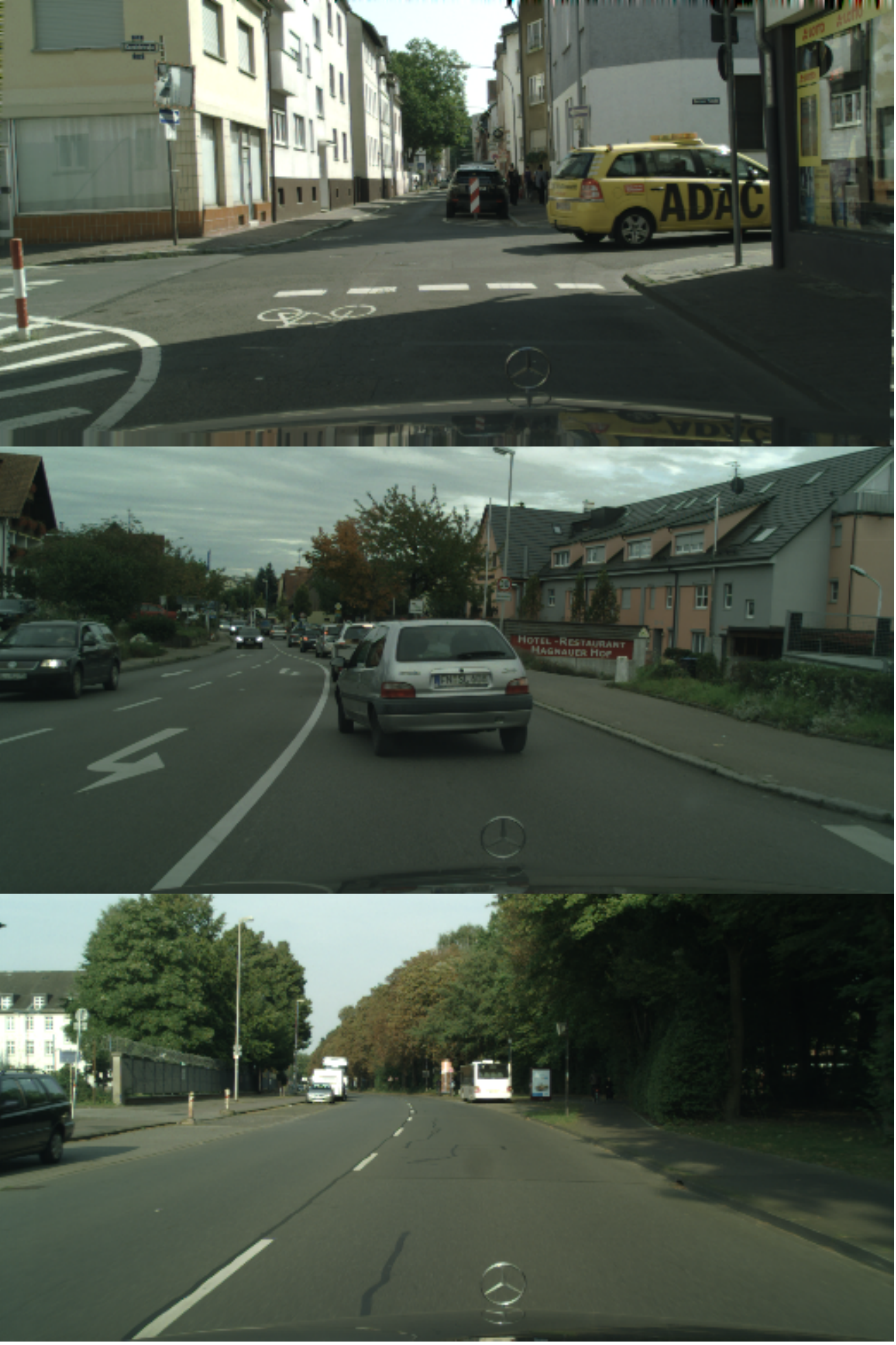}
				\label{fig:city_img}
			\end{minipage}
		} 
		\hspace{-3ex}
		\subfigure[Res18]
		{
			\begin{minipage}{0.18\textwidth}
				\centering
				\includegraphics[width=1.0\linewidth]{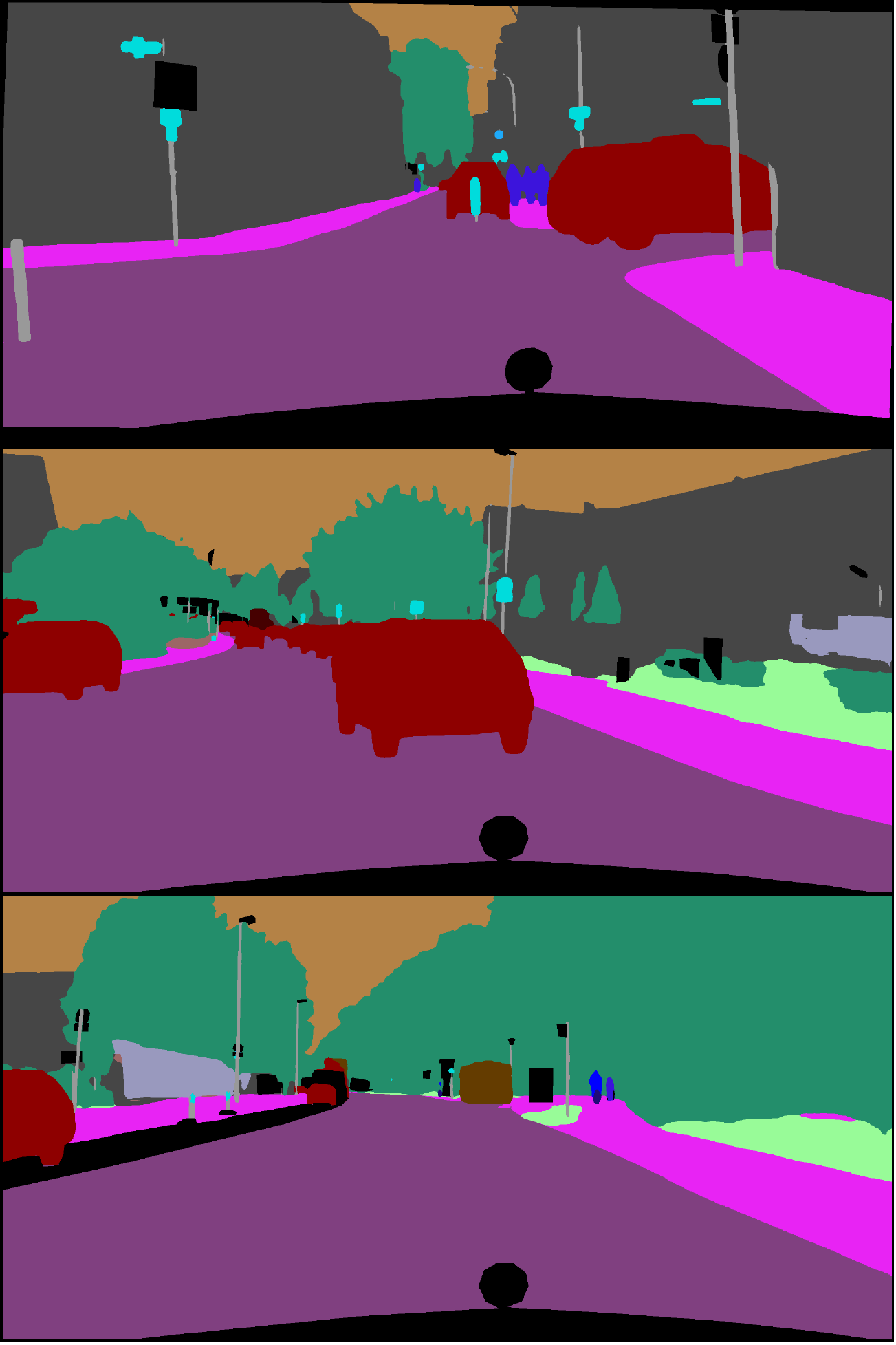}
				\label{fig:city_res18}
			\end{minipage}
		}
		\hspace{-3ex}
		\subfigure[Xception39]
		{
			\begin{minipage}{0.18\textwidth}
				\centering
				\includegraphics[width=1.0\linewidth]{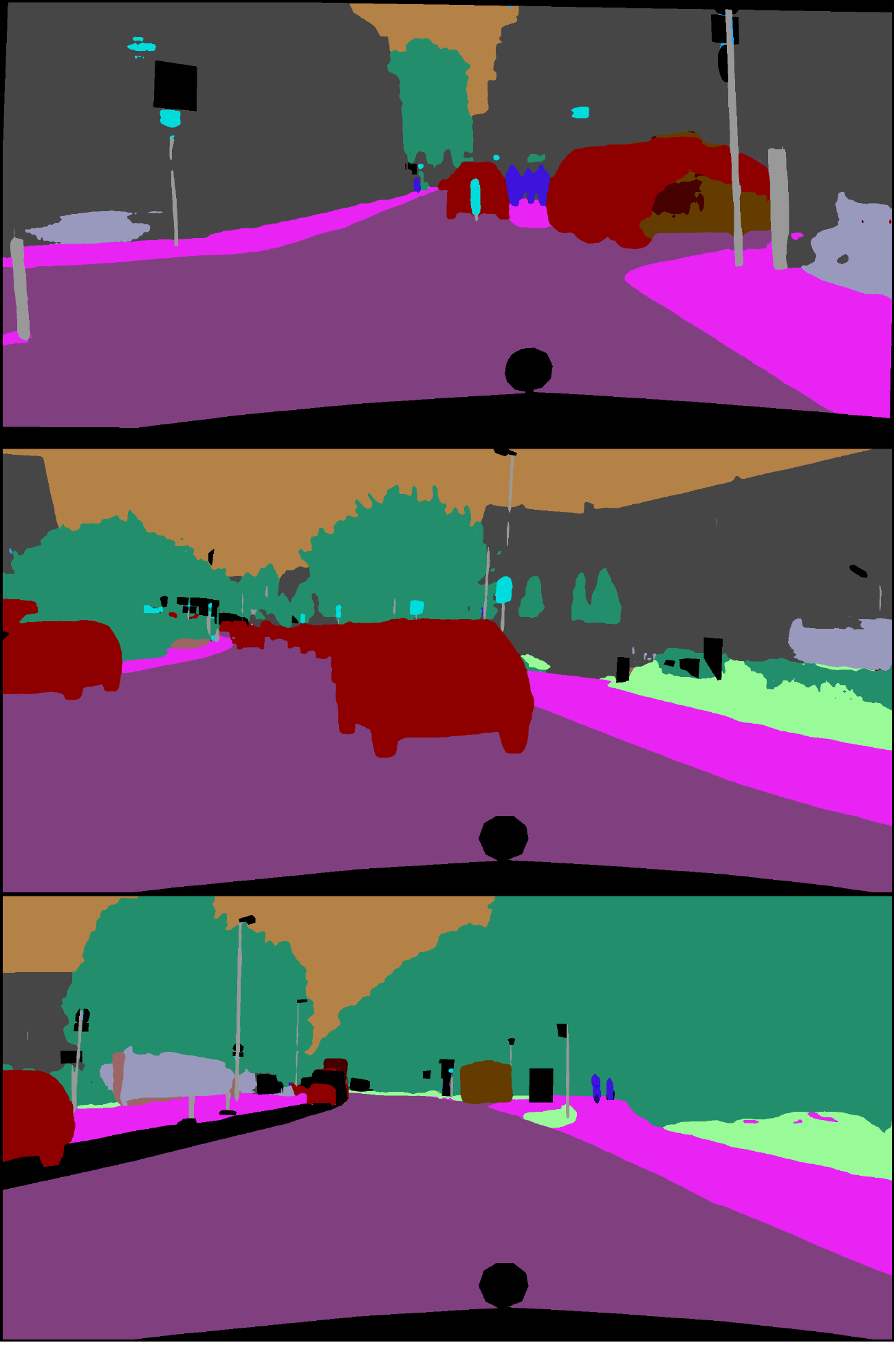}
				\label{fig:city_xception}
			\end{minipage}
		}
		\hspace{-3ex}
		\subfigure[Res101]
		{
			\begin{minipage}{0.18\textwidth}
				\centering
				\includegraphics[width=1.0\linewidth]{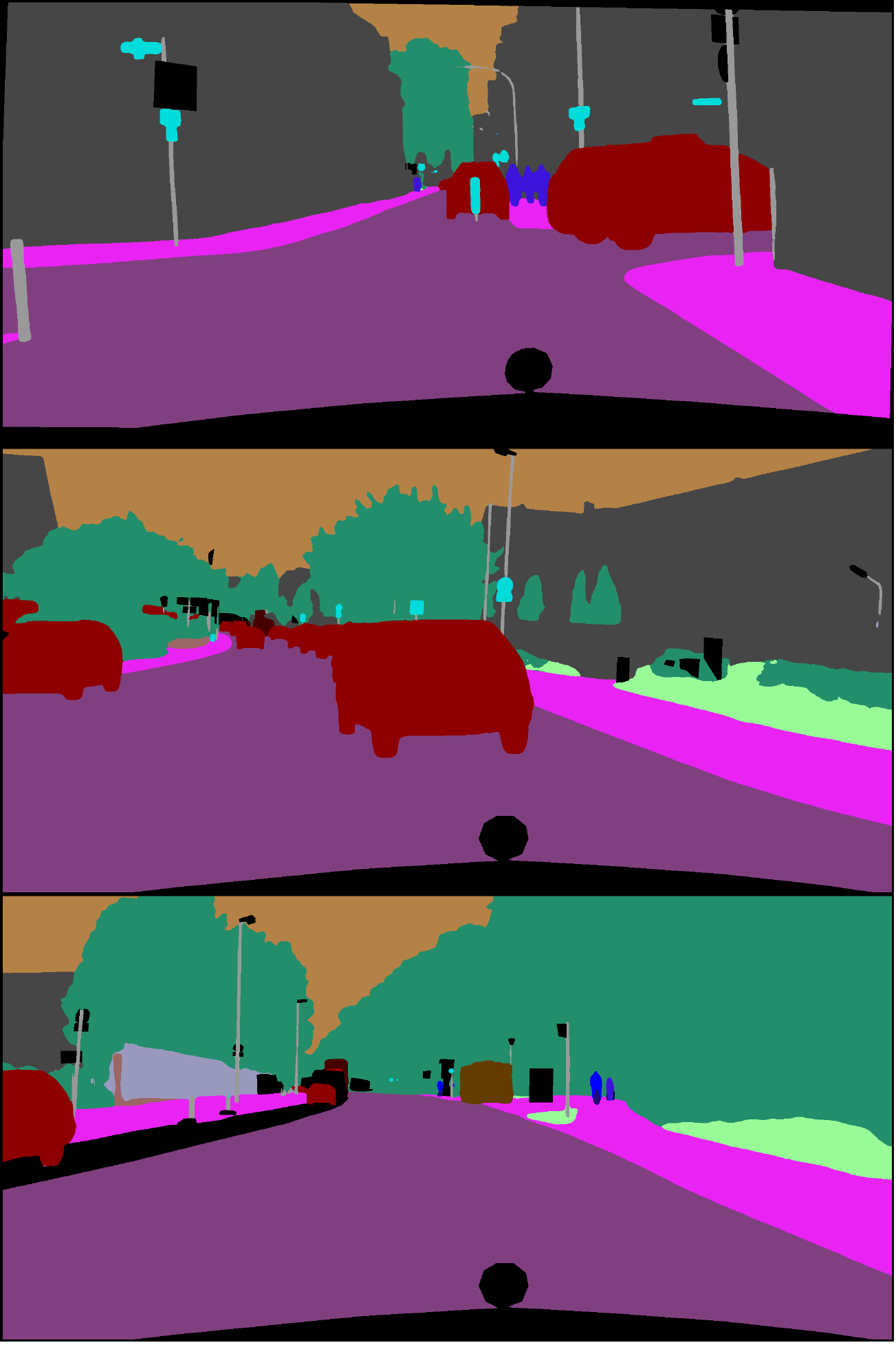}
				\label{fig:city_res101}
			\end{minipage}
		}
		\hspace{-3ex}
		\subfigure[GT]
		{
			\begin{minipage}{0.18\textwidth}
				\centering
				\includegraphics[width=1.0\linewidth]{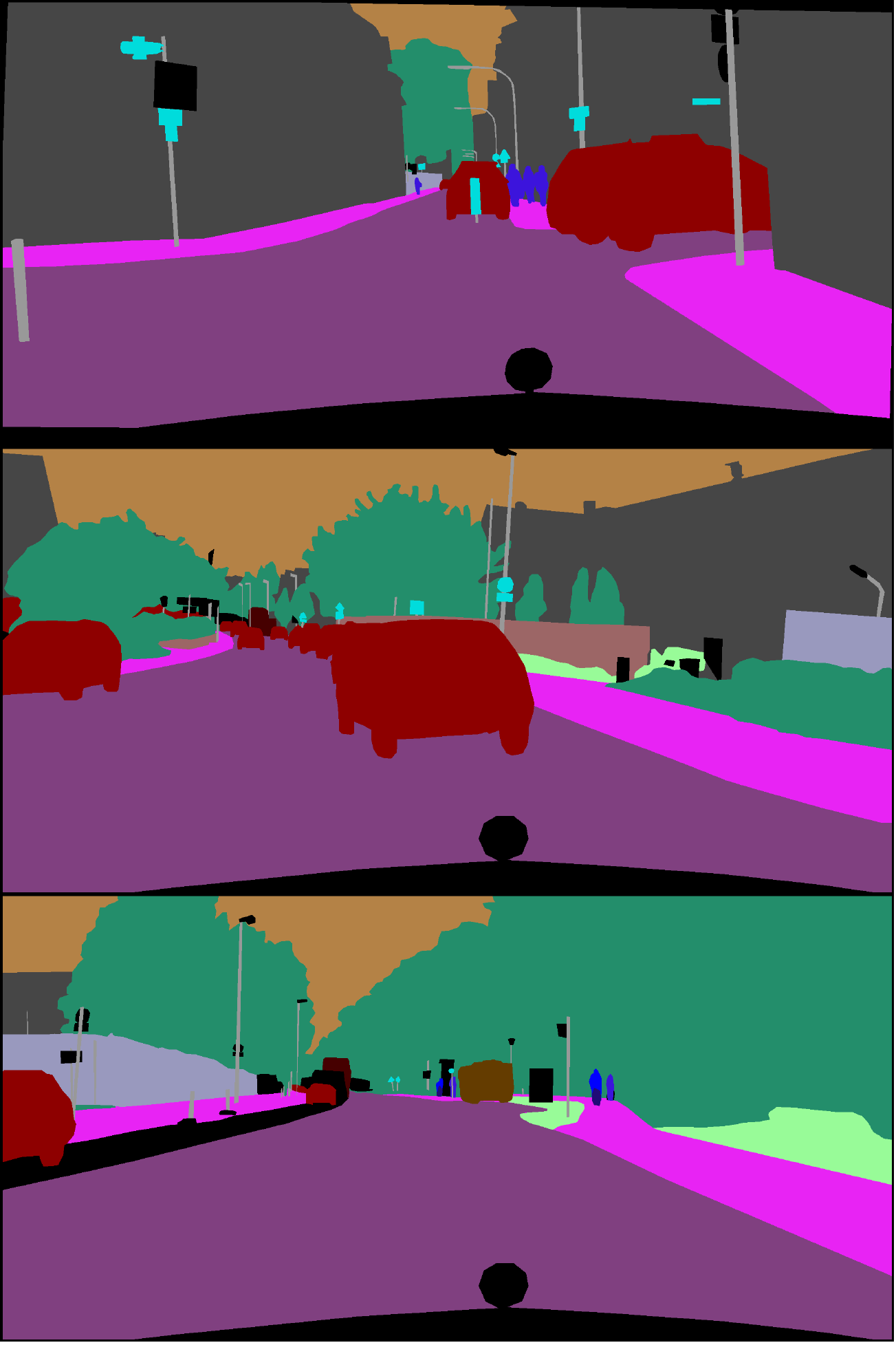}
				\label{fig:city_gt}
			\end{minipage}
		}
		\caption{Example results of the BiSeNet based on Xception39, Res18, and Res101 model on Cityscapes dataset.}
		\label{fig:city_perf}
	\end{figure}
	
\setlength{\tabcolsep}{4pt}
\begin{table}[!t]
	\begin{center}
		\caption{Accuracy comparison of our method against other \emph{state-of-the-art} methods on Cityscapes test dataset.  ``-'' indicates that the methods didn't give the corresponding result.}
		\label{tab:city-performance-comp}
		\begin{tabular}{cccccccc}
			\toprule
			\multicolumn{2}{l}{\multirow{2}*{Method}} & \multicolumn{2}{l}{\multirow{2}*{BaseModel}} & \multicolumn{2}{c}{Mean IOU(\%)} \\
			\cmidrule(lr){5-6}
			\multicolumn{2}{c}{} & \multicolumn{2}{c}{} & \emph{val} & \emph{test} \\
			\noalign{\smallskip}
			\hline
			\noalign{\smallskip}
			\multicolumn{2}{l}{DeepLab~\cite{Chen-ICLR-Deeplabv2-2016}} & \multicolumn{2}{l}{VGG16~\cite{Simonyan-ICLR-VGG-2015}} & - & 63.1  \\
			\multicolumn{2}{l}{FCN-8s~\cite{Long-CVPR-FCN-2015}} & \multicolumn{2}{l}{VGG16} & - & 65.3 \\
			\multicolumn{2}{l}{Adelaide~\cite{Lin-CVPR-Piecewisetraining-2016}} & \multicolumn{2}{l}{VGG16} & - & 66.4  \\
			\multicolumn{2}{l}{Dilation10~\cite{Yu-ICLR-Dilate-2016}} & \multicolumn{2}{l}{VGG16} & 68.7 &  67.1 \\
			\multicolumn{2}{l}{LRR~\cite{Ghiasi-ECCV-LRR-2016}} & \multicolumn{2}{l}{VGG16} & 70.0 &  69.7 \\
			\multicolumn{2}{l}{DeepLab-v2+CRF~\cite{Chen-Arxiv-Deeplabv2-2016}} & \multicolumn{2}{l}{Res101} & 71.4  &  70.4\\
			\multicolumn{2}{l}{RefineNet~\cite{Lin-CVPR-Refinenet-2017}} & \multicolumn{2}{l}{Res101} & - & 73.6 \\
			\multicolumn{2}{l}{DUC~\cite{Wang-CVPR-DUC-2017}} & \multicolumn{2}{l}{Res152} & 76.7 & 76.1 \\
			\multicolumn{2}{l}{PSPNet~\cite{Zhao-CVPR-PSPNet-2017}} & \multicolumn{2}{l}{Res101} & - & \underline{78.4} \\
			\noalign{\smallskip}
			\hline
			\noalign{\smallskip}
			\multicolumn{2}{l}{Ours} & \multicolumn{2}{l}{Xception39} & 72.0 &  71.4\\
			\multicolumn{2}{l}{Ours} & \multicolumn{2}{l}{Res18} & \underline{78.6} & 77.7\\
			\multicolumn{2}{l}{Ours} & \multicolumn{2}{l}{Res101} & \textbf{80.3} & \textbf{78.9}\\
			\bottomrule
		\end{tabular}
	\end{center}
\end{table}
\setlength{\tabcolsep}{1.4pt}

\paragraph{Accuracy analysis:}Actually, our BiSeNet can also achieve higher accuracy result against other non-real-time semantic segmentation algorithms. Here, we will show the accuracy result on Cityscapes~\cite{Cityscapes}, CamVid~\cite{CamVid} and COCO-Stuff~\cite{Caesar-Stuff-2018} benchmarks. Meanwhile, to ensure the validity of our method, we also employ it on different base models, such as the standard ResNet18 and ResNet101~\cite{He-CVPR-ResNet-2016}. Next, we will elaborate on some training details. 
	
	\paragraph{Cityscapes:} 
	As shown in Table~\ref{tab:city-performance-comp}, our method also achieves an impressing result on different models. For improving the accuracy, we take randomly take 1024$\times$1024 crop as input. Here, we only use the fine data of Cityscapes dataset. The Figure~\ref{fig:city_perf} presents some visual examples of our results.
	
	\paragraph{CamVid:} The Table~\ref{tab:result-camvid} shows the statistic accuracy result on CamVid dataset. For testing, we use the training dataset and validation dataset to train our model. Here, we use 960$\times$720 resolution for training and evaluation. 
	
	
	\setlength{\tabcolsep}{3.7pt}
	\begin{table}[!t]
		\begin{center}
			\caption{Accuracy result on CamVid test dataset. \emph{Ours$^1$} and \emph{Ours$^2$} indicate the model based on Xception39 and Res18 network.}
			\label{tab:result-camvid}
			\begin{tabular}{lcccccccccccc}
				\toprule
				Method & \rotatebox{90}{Building} & \rotatebox{90}{Tree} & \rotatebox{90}{Sky} & \rotatebox{90}{Car} & \rotatebox{90}{Sign} & \rotatebox{90}{Road} & \rotatebox{90}{Pedestrain} & \rotatebox{90}{Fence} & \rotatebox{90}{Pole} & \rotatebox{90}{Sidewalk} & \rotatebox{90}{Bicyclist} & \rotatebox{90}{Mean IOU(\%)} \\
				\noalign{\smallskip}
				\hline
				\noalign{\smallskip}
				SegNet-Basic & 75.0 & 84.6 & 91.2 & 82.7 & 36.9 & 93.3 & 55.0 & 47.5 & 44.8 & 74.1 & 16.0 & n/a \\
				SegNet & 88.8 & 87.3 & 92.4 & 82.1 & 20.5 & 97.2 & 57.1 & 49.3 & 27.5 & 84.4 & 30.7 & 55.6\\
				ENet & 74.7 & 77.8 & 95.1 & 82.4 & 51.0 & 95.1 & 67.2 & 51.7 & 35.4 & 86.7 & 34.1 &  51.3 \\
				\noalign{\smallskip}
				\hline
				\noalign{\smallskip}
				Ours$^1$ & 82.2 & 74.4 & 91.9 & 80.8 & 42.8 & 93.3 & 53.8 & 49.7 & 25.4 & 77.3 & 50.0 & \underline{65.6} \\
				Ours$^2$ & 83.0 & 75.8 & 92.0 & 83.7 & 46.5 & 94.6 & 58.8 & 53.6 & 31.9 & 81.4 & 54.0 & \textbf{68.7} \\
				\bottomrule
			\end{tabular}
		\end{center}
	\end{table}
	\setlength{\tabcolsep}{1.4pt}
	
	\paragraph{COCO-Stuff:} We also report our accuracy results on COCO-Stuff validation dataset in Table~\ref{tab:result-stuff}. In the training and validation process, we crop the input into 640$\times$640 resolution. For a fair comparison, we don't adopt the multi-scale testing.
	
	\setlength{\tabcolsep}{3.7pt}
	\begin{table}[!t]
		\begin{center}
			\caption{Accuracy result on COCO-Stuff validation dataset.}
			\label{tab:result-stuff}
			\begin{tabular}{lccc}
				\toprule
				Method & BaseModel & Mean IOU(\%) & Pixel Accuracy(\%) \\
				\noalign{\smallskip}
				\hline
				\noalign{\smallskip}
				Deeplab-v2 & VGG-16 & 24.0 & 58.2 \\
				\noalign{\smallskip}
				\hline
				\noalign{\smallskip}
				Ours & Xception39 & 22.8 &  59.0 \\
				Ours & Res18 & \underline{28.1} & \underline{63.2} \\
				Ours & Res101 & \textbf{31.3} & \textbf{65.5} \\
				\bottomrule
				
			\end{tabular}
		\end{center}
	\end{table}
	\setlength{\tabcolsep}{1.4pt}
	
	
	\section{Conclusions}
	\label{sec:conclusion}
	Bilateral Segmentation Network~(BiSeNet) is proposed in this paper to improve the speed and accuracy of real-time semantic segmentation simultaneously. Our proposed BiSeNet contains two paths: Spatial Path~(SP) and Context Path~(CP). The Spatial Path is designed to preserve the spatial information from original images. And the Context Path utilizes the lightweight model and global average pooling~\cite{Liu-ICLR-ParseNet-2016, Chen-Arxiv-Deeplabv3-2017, Zhao-CVPR-PSPNet-2017} to obtain sizeable receptive field rapidly. With the affluent spatial details and large receptive field, we achieve the result of 68.4\% Mean IOU on Cityscapes~\cite{Cityscapes} test dataset at 105 FPS. 
	
	\section*{Acknowledgment}
	This work was supported by the Project of the National Natural Science Foundation of China No.61433007 and No.61401170.

%
%
%
\bibliographystyle{splncs04}
\bibliography{egbib}
%




\end{document}

%% file: macro.tex
\usepackage{epsfig}
\usepackage{xspace}
\usepackage{color}
\usepackage{multirow}
\usepackage{graphics}
\usepackage{adjustbox}
\usepackage{subfigure}
\usepackage{amsmath}
\usepackage[noend]{algpseudocode}
\usepackage{algorithmicx,algorithm}
\usepackage{paralist} 
\usepackage{enumitem}
\usepackage{graphicx} 
\usepackage{epstopdf}
\usepackage{booktabs} 

\usepackage[pagebackref=true,breaklinks=true,letterpaper=true,colorlinks,citecolor=blue,linkcolor=blue,bookmarks=false]{hyperref}

\long\def\ignorethis#1{}

\definecolor{Gray}{rgb}{0.35,0.35,0.35}
\definecolor{Blue}{rgb}{0,0.2,0.8}
\definecolor{Red}{rgb}{0.8,0.2,0}
\definecolor{Green}{rgb}{0.0,0.5,0.1}
\definecolor{Gray}{rgb}{0.4,0.4,0.4}

\newlength\paramargin
\newlength\figmargin
\newlength\secmargin

\usepackage{array}
\newcolumntype{L}[1]{>{\raggedright\let\newline\\\arraybackslash\hspace{0pt}}m{#1}}
\newcolumntype{C}[1]{>{\centering\let\newline\\\arraybackslash\hspace{0pt}}m{#1}}
\newcolumntype{R}[1]{>{\raggedleft\let\newline\\\arraybackslash\hspace{0pt}}m{#1}}

\graphicspath{{figure/eps/}}